\documentclass[letterpaper, 10 pt, conference]{ieeeconf}  

\IEEEoverridecommandlockouts                              

\usepackage{amsmath} 
\usepackage{amssymb}  

\usepackage{graphicx}
\usepackage{caption}
\usepackage{booktabs}
\usepackage{float}
\usepackage{cite}
\usepackage{color}
\usepackage[table]{xcolor}
\usepackage{multirow}
\usepackage{fontawesome5}
\usepackage{algorithm}
\usepackage{algpseudocode}
\usepackage{gensymb}
\usepackage{hyperref}
\usepackage{fancyhdr}
\usepackage{xspace}
\usepackage{pifont}

\hypersetup{colorlinks=true, linkcolor=blue!50!black,
            citecolor=blue!50!black, urlcolor=blue!50!black}
            
\title{\textbf{NeSAM: Neuro-Symbolic Kinodynamics with\\ Soil Adaptation for Off-Road Mobility
}}

\author{
Chenhui Pan$^{1}$, Tong Xu$^{1}$, Francesco Cancelliere$^{1,2}$, and Xuesu Xiao$^{1}$%
\thanks{$^{1}$Chenhui Pan, Tong Xu, Francesco Cancelliere and Xuesu Xiao are with the Department of Computer Science, George Mason University, USA.}%
\thanks{$^{2}$Francesco Cancelliere is also with the Department of Electrical, Electronic and Computer Engineering, University of Catania, Italy.}%
}

\begin{document}

\maketitle
\thispagestyle{empty}
\pagestyle{empty}

\begin{abstract}
Accurate prediction of off-road vehicle motion over deformable terrain remains challenging because sinkage, slip, and traction vary with local soil conditions. Existing learning-based kinodynamic models directly approximate vehicle-terrain interactions from data but do not explicitly represent soil mechanics and offer limited physical interpretability. To address these limitations, we present \textit{NeSAM}, a neuro-symbolic framework that combines differentiable Bekker-Wong terramechanics with learned terrain representations and a Transformer-based residual dynamics model for long-horizon, six degree-of-freedom kinodynamic prediction. The terramechanics component models soil-dependent interaction forces, while the residual model corrects discrepancies between the analytical prediction and the observed vehicle dynamics. NeSAM further estimates physically meaningful soil parameters from terrain observations and updates them online using an extended Kalman filter. We evaluate NeSAM in Verti-Bench, a simulator built on the Chrono multiphysics engine, and validate its performance on a physical Verti-4-Wheeler platform. NeSAM improves prediction accuracy by up to \(30\%\) in simulation and \(29\%\) on real-world data relative to the strongest compared baselines. When integrated with a close-loop navigation controller, NeSAM further improves traversal success rate through online soil adaptation while reduces Hausdorff distance to the reference trajectory by \(69.4\%\), indicating improved trajectory tracking accuracy.
\end{abstract}
\section{Introduction}
\label{sec:introduction}

Autonomous ground vehicles operating in off-road environments must predict how terrain geometry and material properties affect their motion. On deformable, uneven off-road terrain, pressure-dependent sinkage, shear deformation, and wheel slip cause the realized vehicle motion to differ from that predicted by conventional planar kinematic models\cite{bekker1969,janosi1961,wong2008}. These effects influence not only vehicle translation but also vehicle attitude. Consequently, mobility over rugged and deformable terrain requires a deformation-conditioned, six degree-of-freedom kinodynamic model that can be repeatedly queried during predictive planning\cite{datar2024toward, datar2024learning,datar2024terrain,vertibench2025}.

Classical terramechanics models directly describe deformable wheel-terrain interactions through analytical relationships. In particular, the Bekker pressure-sinkage model and the Janosi-Hanamoto shear model characterize soil response using physically interpretable parameters related to stiffness, cohesion internal friction, and shear deformation~\cite{bekker1969,janosi1961,wong2008,krenn2009scm}. This physical structure supports explicit reasoning about traction and soil deformation. However, these models rely on prescribed constitutive relationships that may not capture suspension response, tire deformation, and other vehicle-dependent effects. Their soil parameters are also difficult to estimate and may vary as the vehicle traverses heterogeneous terrain. Although optimized soil-contact simulators can operate in real time, their computational cost remains challenging for predictive controllers that require many sequential kinodynamics evaluations within each control cycle~\cite{serban2023scm,datar2024terrain}.

\begin{figure}[t!]
    \centering
    \includegraphics[width=\columnwidth]{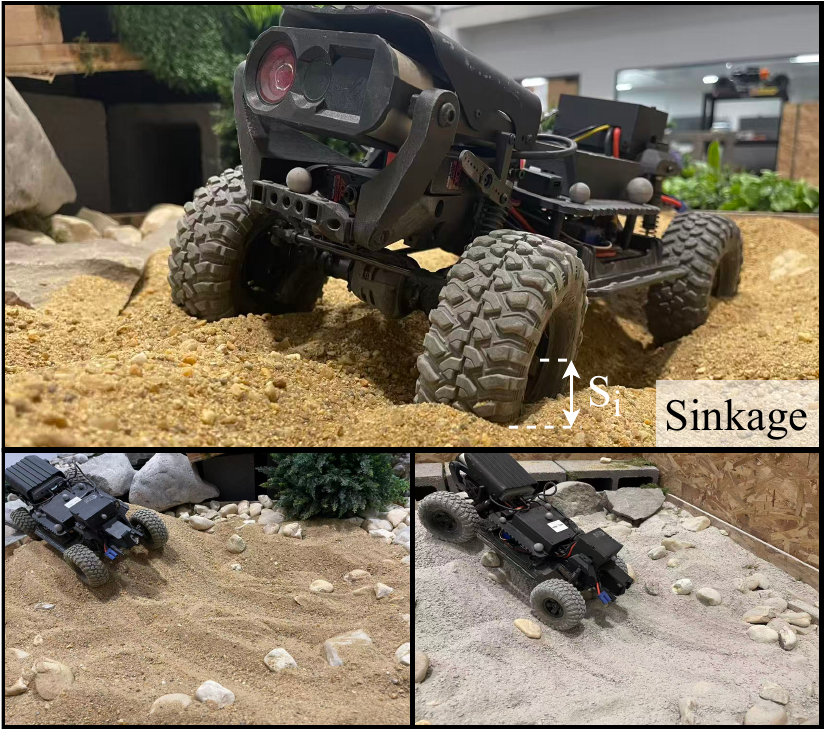}
    \caption{Motivation for NeSAM: deformable terrain causes wheel sinkage
    and persistent terrain deformation, motivating deformation-aware
    kinodynamic modeling and online soil adaptation.}
    \label{fig::motivation}
    \vspace{-5pt}
\end{figure}

Learning-based kinodynamics models approximate nonlinear vehicle-terrain interactions directly from vehicle states, control
inputs, and terrain observations ~\cite{triest2022tartandrive,xiao2021learning,karnan2022,pokhrel2024cahsor}. Their data-driven formulation can capture behaviors that are difficult to represent using fixed analytical assumptions. Recent methods further improve query efficiency through state space decomposition~\cite{datar2024learning} or self-supervised representation learning~\cite{datar2024terrain}. Physics-informed models combine learned and analytical traversability estimates under distribution shift~\cite{pietra2025}. However, these methods generally treat terrain as a rigid surface, without modeling its deformation under various vehicle load distribution or exposing interpretable soil parameters. Capturing such effects entirely from data would require broad coverage of soil, loading, state, and control conditions, which becomes difficult in diverse off-road conditions. 

These limitations motivate NeSAM (\textbf{Ne}uro-symbolic \textbf{S}oil \textbf{A}daptive \textbf{M}obility), which combines learned terrain representations, differentiable Bekker-Wong terramechanics, and a Transformer-based residual model. The terramechanics component provides deformable-soil structure and adaptable physical parameters, while the learned residual accounts for systematic discrepancies between the analytical prediction and observed vehicle motion.
The main contributions of this work are summarized as follows:
\begin{itemize}
    \item We present a neuro-symbolic kinodynamics framework that
    integrates differentiable Bekker--Wong terramechanics, learned
    terrain representations, and Transformer-based residual kinodynamics
    for long-horizon, six degree-of-freedom vehicle motion prediction
    over deformable terrain;

    \item We develop an EKF-based online soil adaptation method that
    updates physically interpretable terramechanics parameters using
    differences between predicted and measured vehicle motion;

    \item We evaluate long-horizon autoregressive prediction in
    Chrono-based simulation and on real-world vehicle trajectories,
    with comparisons against terrain-aware and terrain-free
    learning-based models; and

    \item We validate online soil adaptation through closed-loop
    navigation in both simulation and on a physical Verti-4-Wheeler platform.
\end{itemize}

\section{Related Work}
\label{sec:related_work}

In this section, we review prior work on terramechanics and soil identification, learning-based off-road mobility modeling, and online adaptation of vehicle kinodynamics.

\subsection{Terramechanics and Soil Identification}

Foundational terramechanics models describe normal and tangential wheel-soil forces using the Bekker pressure-sinkage relationship and the Janosi-Hanamoto shear formulation~\cite{bekker1969,janosi1961,wong2008}. These models have been incorporated into multibody simulation frameworks, including the Soil Contact Model (SCM), to reproduce vehicle motion on deformable terrain~\cite{krenn2009scm,tasora2015chrono}. Subsequent work has focused on reducing their computational cost and identifying uncertain soil parameters. Dallas \textit{et al.} introduced a nonlinear SCM surrogate with unscented-Kalman-filter estimation of the soil sinkage exponent~\cite{dallas2020online}, and later developed a differentiable neural approximation for force prediction and terrain-parameter estimation~\cite{dallas2020neural}. These studies primarily address wheel-level force modeling, rather than full-state vehicle prediction within a planning loop.

\subsection{Learning-based Off-Road Kinodynamics}

Learning-based off-road mobility has progressed from inverse models and end-to-end control policies to forward kinodynamics models for predictive planning. Inverse kinodynamic models learn control corrections from vehicle motion data~\cite{xiao2021learning,karnan2022}, while imitation learning directly maps onboard observations to steering and throttle commands~\cite{pan2018agile}. In contrast, learned forward models predict future vehicle states and can therefore be incorporated into model-predictive controllers~\cite{atreya2022high}.

Recent work has increasingly conditioned forward kinodynamics on terrain observations. TartanDrive provides multimodal vehicle and terrain data for learning off-road kinodynamics~\cite{triest2022tartandrive}. Lee \textit{et al.} combined proprioceptive and exteroceptive inputs to predict six-degree-of-freedom vehicle motion and use predictive uncertainty during Model Predictive Path Integral (MPPI) planning~\cite{lee2023terrain}. Gibson \textit{et al.} used visual terrain features to model changes in high-speed vehicle kinodynamics across different terrain~\cite{gibson2025visual}. For vertically challenging terrain, Datar \textit{et al.} employed a decomposed six degree-of-freedom model~\cite{datar2024learning}, whereas Terrain-Attentive Learning leverages self-supervised learning~\cite{datar2024terrain}, both to improve real-time model query efficiency for repeated use within a predictive planner.

These methods differ in their state representations, terrain modalities, and prediction structures, but they all treat terrain as rigid surfaces and terrain-dependent effects are primarily encoded through features learned from scratch. They do not explicitly consider the load-dependent soil variables governing sinkage and shear on deformable terrain or provide corresponding physical parameters for online soil adaptation.

\subsection{Online Adaptation and Hybrid Kinodynamics}

Online adaptation methods update a kinodynamics model using recent observations through meta-learning, latent-state inference, or learned adaptation policies~\cite{clavera2018rl,nagabandi2019deepmeta,kumar2021rma}. In off-road mobility, Kalman-filter-based updates have been used to adapt learned vehicle models during deployment~\cite{levy2025meta}. VertiAdaptor represents terrain-conditioned kinodynamics with neural ordinary differential equation basis functions and adapts their coefficients through a least-squares update using recent trajectory data~\cite{vertiadaptor2026}. These approaches modify learned parameters or latent coefficients rather than explicit soil-mechanics variables.

Differentiable simulators provide another route for integrating physics with learning and have been applied to system identification, control, and residual modeling in rigid- and soft-body systems~\cite{degrave2019,hu2019chainqueen,heiden2021dispn,murthy2021gradsim}. Existing work, however, has not jointly addressed terrain-conditioned six degree-of-freedom prediction, differentiable deformable-soil mechanics, and adaptation of interpretable soil parameters within an off-road planning framework.
\section{Method}
\label{sec:method}

\begin{figure*}[t!]
    \centering
    \includegraphics[width=\textwidth]{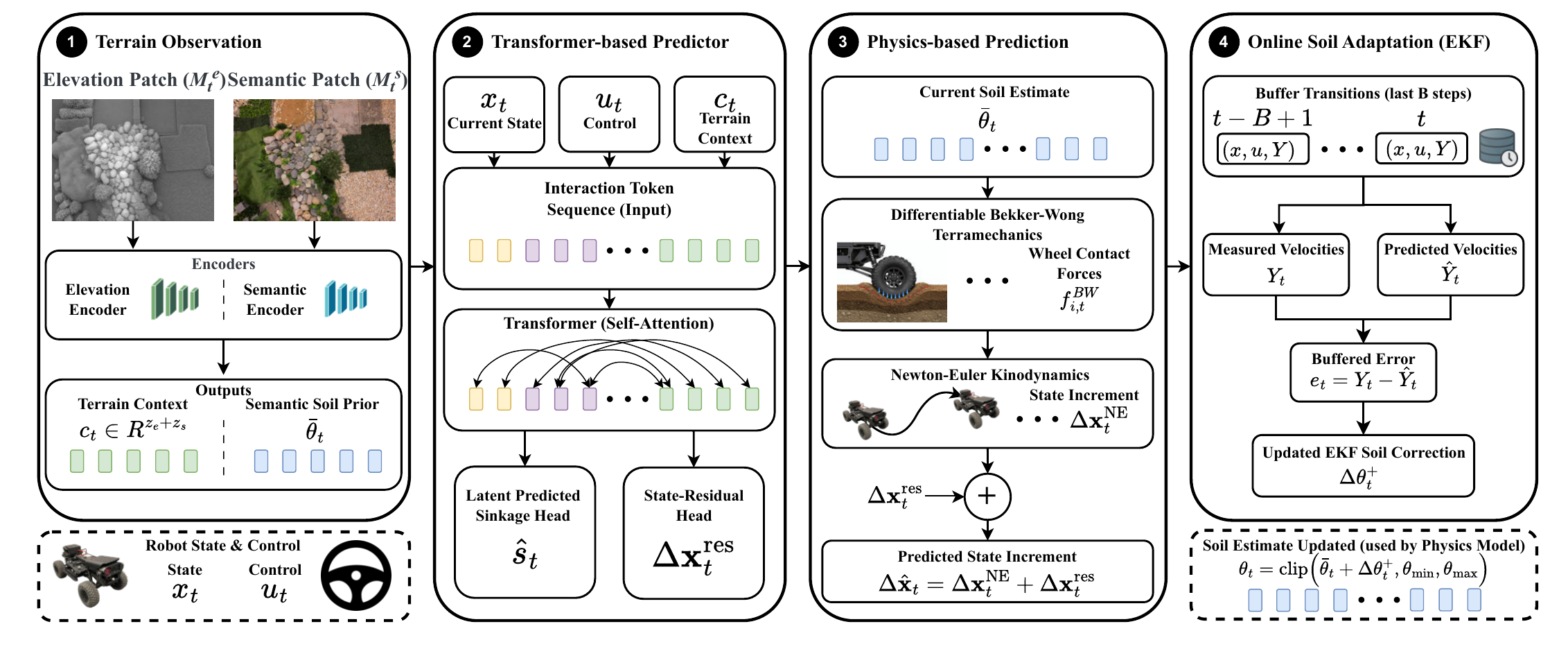}
    \caption{Overview of NeSAM. Elevation and semantic observations
    \((\mathbf{m}^{\mathrm e}_t,\mathbf{m}^{\mathrm s}_t)\) are encoded to
    obtain the terrain context \(\mathbf{c}_t\) and nominal soil estimate
    \(\bar{\boldsymbol{\theta}}_t\). Together with the vehicle state
    \(\mathbf{x}_t\) and control \(\mathbf{u}_t\), the terrain context forms
    the interaction-token history processed by the Transformer. The
    sinkage head predicts wheel sinkages \(\hat{\mathbf{s}}_t\)
    together with the soil estimate, are used by the
    differentiable Bekker-Wong model to compute wheel-terrain forces
    \(\mathbf{f}^{\mathrm{BW}}_{i,t}\). Newton-Euler kinodynamics converts
    these forces into the physics-based state increment
    \(\Delta\mathbf{x}^{\mathrm{NE}}_t\), which is combined with the learned
    state residual \(\Delta\mathbf{x}^{\mathrm{res}}_t\) to predict the
    next vehicle state. During deployment, measured motion updates the soil
    correction \(\Delta\boldsymbol{\theta}\) through the buffered EKF, while
    all learned model components remain fixed.}
    \label{fig:nesam_framework}
\vspace{-5pt}
\end{figure*}

NeSAM predicts vehicle motion by coupling learned terrain-conditioned
interaction modeling with differentiable terramechanics. As illustrated
in Fig.~\ref{fig:nesam_framework}, a Transformer encodes the recent
vehicle-terrain interaction history. A sinkage head predicts the
wheel sinkages required by the terramechanics model, while a
state-residual head predicts a single chassis-level correction to the
physics-based state increment. The predicted sinkages, together with
the wheel angular velocities available to the model and the current
soil estimate, determine the wheel-terrain interaction forces through
differentiable terramechanics. These forces are propagated through
Newton-Euler kinodynamics, and the learned state residual corrects the
resulting vehicle-state prediction.

\subsection{Problem Formulation}
\label{sec:problem_formulation}

We consider a ground vehicle operating over uneven, deformable terrain.
At time \(t\), the vehicle state is
\[
    \mathbf{x}_t
    =
    \left[
        \mathbf{p}_t,
        \boldsymbol{\eta}_t,
        \mathbf{v}_t,
        \boldsymbol{\omega}_t
    \right]
    \in\mathbb{R}^{12}.
\]
Here, \(\mathbf{p}_t\in\mathbb{R}^{3}\) is the vehicle position,
\(\boldsymbol{\eta}_t\in\mathbb{R}^{3}\) contains its roll, pitch, and
yaw angles, and \(\mathbf{v}_t,\boldsymbol{\omega}_t\in\mathbb{R}^{3}\)
are its linear and angular velocities, respectively. The control input is
\(\mathbf{u}_t\in\mathcal{U}\) corresponds to speed and steering
angle.

The local terrain observation is
\(\mathbf{m}_t=(\mathbf{m}^{\mathrm e}_t,
\mathbf{m}^{\mathrm s}_t)\), where
\(\mathbf{m}^{\mathrm e}_t\) is a vehicle-aligned elevation patch and
\(\mathbf{m}^{\mathrm s}_t\) is the corresponding semantic terrain
patch. The elevation and semantic observations are processed by the
learned encoders \(E_{\mathrm e}\) and \(E_{\mathrm s}\), respectively,
producing latent representations
\(\mathbf{z}^{\mathrm e}_t=E_{\mathrm e}
(\mathbf{m}^{\mathrm e}_t)\) and
\(\mathbf{z}^{\mathrm s}_t=E_{\mathrm s}
(\mathbf{m}^{\mathrm s}_t)\).

We directly concatenate
the two representations to form the terrain context
\(\mathbf{c}_t=
[\mathbf{z}^{\mathrm e}_t\mathbin{\|}
\mathbf{z}^{\mathrm s}_t]\), where \(\mathbin{\|}\) denotes feature
concatenation. The terrain context conditions the Transformer-based
vehicle-terrain interaction model. In parallel, a soil-prior head \(h_{\theta}\) maps the semantic
representation to the nominal soil parameters,
\(\bar{\boldsymbol{\theta}}_t
=h_{\theta}(\mathbf{z}^{\mathrm s}_t)\). This semantic estimate provides
the initial soil parameters used by the differentiable terramechanics
model and is subsequently refined during deployment by the online
adaptation procedure described in Sec.~\ref{sec:online_soil_adaptation}.

We define the complete one-step terrain-conditioned transition as
\begin{equation}
    \hat{\mathbf{x}}_{t+1}
    =
    F_{\boldsymbol{\Phi}}\!\left(
        \mathbf{x}_t,
        \mathbf{u}_t,
        \mathbf{m}_t;
        \bar{\boldsymbol{\theta}}_t
    \right),
    \label{eq:forward_kinodynamics}
\end{equation}
where \(\hat{\mathbf{x}}_{t+1}\) is the predicted vehicle state at the
next time step, \(\boldsymbol{\Phi}\) describes all learned model
parameters, and \(\bar{\boldsymbol{\theta}}_t\) denotes the soil parameters
currently used by the differentiable terramechanics model. 

The transition \(F_{\boldsymbol{\Phi}}\) comprises terrain encoding,
Transformer-based wheel-variable and state-residual prediction,
differentiable terramechanics, and Newton-Euler propagation. Given an
initial state \(\mathbf{x}_t\) and a future control sequence
\(\mathbf{u}_{t:t+K-1}\), the transition is recursively applied for
\(K\) steps to predict the vehicle trajectory over the planning
horizon.

\subsection{Neuro-Symbolic Wheel-Terrain Interaction}
\label{sec:neuro_symbolic_interaction}

Given the recent vehicle states, controls, and terrain contexts, we
construct the interaction-token sequence
\[
    \boldsymbol{\zeta}_{j}
    =
    \mathbf{x}_{j}
    \mathbin{\|}
    \mathbf{u}_{j}
    \mathbin{\|}
    \mathbf{c}_{j},
    \qquad
    j=t-L+1,\ldots,t,
\]
where \(L\) is the history length. The Transformer encodes this sequence
as
\[
    \mathbf{h}_t
    =
    T_{\boldsymbol{\psi}}
    \left(
        \boldsymbol{\zeta}_{t-L+1:t}
    \right).
\]
The sinkage head maps the Transformer representation to the wheel
sinkages required by the terramechanics model,
\[
\hat{\mathbf{s}}_t
=
H_{\mathrm{sink}}(\mathbf{h}_t),
\]
where
\(\hat{\mathbf{s}}_t=
[\hat{s}_{1,t},\ldots,\hat{s}_{N_w,t}]\).
Thus, terrain-dependent wheel sinkage is inferred from the recent
vehicle-terrain interaction history rather than from a single terrain
observation. In parallel, the state-residual head predicts the
chassis-level correction
\(\Delta\mathbf{x}^{\mathrm{res}}_t
=H_{\mathrm{res}}(\mathbf{h}_t)\).

For each wheel \(i\), the wheel angular velocity
\(\Omega_{i,t}\), predicted sinkage \(\hat{s}_{i,t}\), current vehicle
state, and soil estimate instantiate the differentiable Bekker-Wong
model:
\[
\mathbf{f}^{\mathrm{BW}}_{i,t}
=
\mathcal{B}
\left(
\Omega_{i,t},
\hat{s}_{i,t},
\mathbf{x}_t,
\mathbf{m}^{\mathrm e}_t;
\bar{\boldsymbol{\theta}}_t
\right).
\]
Here, \(\mathbf{f}^{\mathrm{BW}}_{i,t}\in\mathbb{R}^{3}\) denotes the
resultant wheel-terrain contact force at wheel \(i\), and
\(\mathcal{B}(\cdot)\) denotes the differentiable pressure-sinkage and
shear-force calculations. Unlike the sinkage
\(\hat{s}_{i,t}\), the wheel angular velocity \(\Omega_{i,t}\) is not
predicted by the Transformer. These contact forces serve as inputs to the
Newton-Euler kinodynamics model described in
Sec.~\ref{sec:state_increment}. Thus, the wheel branch predicts the
physical variables required by the terramechanics model rather than a
learned force residual.

\subsection{Newton-Euler Kinodynamics Model Prediction}
\label{sec:state_increment}

The terramechanics model provides the contact force
\(\mathbf{f}^{\mathrm{BW}}_{i,t}\) at each wheel. These forces are
aggregated to obtain the net force and moment acting on the vehicle:
\[
    \mathbf{F}_t
    =
    \sum_{i=1}^{N_w}
    \mathbf{f}^{\mathrm{BW}}_{i,t}
    +
    M\mathbf{g},
    \qquad
    \boldsymbol{\tau}_t
    =
    \sum_{i=1}^{N_w}
    \mathbf{r}_{i,t}
    \times
    \mathbf{f}^{\mathrm{BW}}_{i,t},
\]
where \(N_w\) is the number of wheels, \(M\) is the vehicle mass,
\(\mathbf{g}\) is gravitational acceleration, and
\(\mathbf{r}_{i,t}\) is the position of wheel \(i\) relative to the
vehicle center of mass.

The corresponding linear and angular accelerations follow the
Newton-Euler equations:
\[
    \mathbf{a}_t
    =
    \frac{\mathbf{F}_t}{M},
    \qquad
    \dot{\boldsymbol{\omega}}_t
    =
    \mathbf{I}_t^{-1}
    \left(
        \boldsymbol{\tau}_t
        -
        \boldsymbol{\omega}_t
        \times
        \mathbf{I}_t\boldsymbol{\omega}_t
    \right),
\]
where \(\mathbf{I}_t\) denotes the vehicle inertia tensor. Numerical
integration over the sampling interval \(\Delta t\) converts these
accelerations into the physics-based state increment
\(\Delta\mathbf{x}^{\mathrm{NE}}_t\), including changes in position,
orientation, linear velocity, and angular velocity.

The Transformer-based state residual
\(\Delta\mathbf{x}^{\mathrm{res}}_t\), defined in
Sec.~\ref{sec:neuro_symbolic_interaction}, compensates for
vehicle-level state-transition errors that are not explicitly
represented by the analytical model. Such discrepancies may arise from
chassis and suspension compliance, load-transfer effects, simplified
wheel-contact geometry, and numerical integration error. The complete NeSAM state update is therefore
\begin{equation}
    \hat{\mathbf{x}}_{t+1}
    =
    \mathbf{x}_t
    +
    \Delta\mathbf{x}^{\mathrm{NE}}_t
    +
    \Delta\mathbf{x}^{\mathrm{res}}_t.
    \label{eq:nesam_state_update}
\end{equation}
Equation~\eqref{eq:nesam_state_update} instantiates the complete
transition \(F_{\boldsymbol{\Phi}}(\cdot)\) introduced in
Eq.~\eqref{eq:forward_kinodynamics}. The Newton-Euler component
provides the physically structured motion estimate, while the learned
state residual compensates for the remaining state-transition error.

\subsection{NeSAM Learning and Adaptation}
\label{sec:nesam_learning}

\subsubsection{Offline Autoregressive Training}

The learned components of NeSAM are optimized through
multi-step autoregressive prediction. Starting from
\(\hat{\mathbf{x}}_t=\mathbf{x}_t\), the rollout is generated for
\(k=0,\ldots,K-1\) by recursively applying
\[
    \hat{\mathbf{x}}_{t+k+1}
    =
    F_{\boldsymbol{\Phi}}\!\left(
        \hat{\mathbf{x}}_{t+k},
        \mathbf{u}_{t+k},
        \hat{\mathbf{m}}_{t+k};
        \bar{\boldsymbol{\theta}}_{t+k}
    \right).
\]
Here, \(\hat{\mathbf{m}}_{t+k}\) is the terrain observation aligned with
the predicted vehicle state, and
\(\bar{\boldsymbol{\theta}}_{t+k}\) is the corresponding semantic soil
prior. Thus, each predicted state is used to construct the subsequent
model input without access to the future ground-truth state.
The model is trained using a multi-step rollout objective
\begin{equation}
    \mathcal{L}
    =
    \sum_{k=1}^{K}
    \lambda_k
    \ell\!\left(
        \hat{\mathbf{x}}_{t+k},
        \mathbf{x}_{t+k}
    \right),
    \label{eq:rollout_loss}
\end{equation}
where \(\ell(\cdot,\cdot)\) measures the weighted vehicle-state
prediction error and \(\lambda_k\) controls the contribution of each
rollout step. Gradients are propagated through the soil predictor, Transformer
prediction heads, differentiable terramechanics, and Newton-Euler
kinodynamics model, while the pretrained terrain encoders remain fixed.

\subsubsection{Online Soil Adaptation}
\label{sec:online_soil_adaptation}

During deployment, all learned components of NeSAM remain
fixed. Only the soil parameters used by the differentiable
terramechanics model are adapted. For each terrain observation, the
semantic soil-prior head provides the nominal estimate
\[
    \bar{\boldsymbol{\theta}}_k
    =
    h_{\theta}\!\left(
        E_{\mathrm{s}}(\mathbf{m}^{\mathrm{s}}_k)
    \right),
\]
which is used directly before an online correction is available.

Let \(t\) denote the beginning of a non-overlapping buffer of \(B\)
transitions. The latest posterior soil correction
\(\Delta\boldsymbol{\theta}^{+}_{t}\) and covariance
\(\mathbf{P}^{+}_{t}\) remain fixed while the buffer is collected. For
each \(k=t,\ldots,t+B-1\), the soil estimate used by the
terramechanics model is
\[
    \boldsymbol{\theta}_k
    =
    \operatorname{clip}\!\left(
        \bar{\boldsymbol{\theta}}_k
        +
        \Delta\boldsymbol{\theta}^{+}_{t},
        \boldsymbol{\theta}_{\min},
        \boldsymbol{\theta}_{\max}
    \right),
\]
where the clipping operation enforces physically admissible parameter
bounds. We initialize
\(\Delta\boldsymbol{\theta}^{+}_{0}=\mathbf{0}\), so the initial soil
estimate is given entirely by the semantic prior. The resulting
parameters are used by the complete NeSAM transition to predict
the next linear and angular velocities
(lines~\ref{line:collect_start}-\ref{line:collect_end}).

After \(B\) transitions are collected, the predicted and measured
velocities are stacked into \(\hat{\mathbf{Y}}_t\) and
\(\mathbf{Y}_t\), respectively
(lines~\ref{line:stack_pred}-\ref{line:end_pred}). Their difference,
together with the sensitivity of the predicted motion to the soil
parameters, is used to perform one extended Kalman filter (EKF) update
(lines~\ref{line:stack_buffer}-\ref{line:ekf_update}). The resulting
posterior correction
\(\Delta\boldsymbol{\theta}^{+}_{t+B}\) is then retained while the next
non-overlapping buffer is collected beginning at \(t+B\)
(line~\ref{line:restart_buffer}). Thus, the semantic model always
provides the nominal soil estimate, while measured vehicle motion
periodically refines it through the EKF. The complete procedure is
summarized in Algorithm~\ref{alg:online_soil_adaptation}.

\begin{algorithm}[t]
    \footnotesize
    \caption{Online Soil Adaptation via EKF}
    \label{alg:online_soil_adaptation}
    \begin{algorithmic}[1]
        \Require Fixed transition model \(F_{\boldsymbol{\Phi}}\);
        semantic soil-prior mapping \(h_{\theta}\), \(E_{\mathrm{s}}\)
        \Require Velocity extractor
        \(\mathcal{V}(\mathbf{x})
        =
        [\mathbf{v},\boldsymbol{\omega}]\)
        \Require Buffer size \(B\); admissible soil bounds
        \(\boldsymbol{\theta}_{\min},
        \boldsymbol{\theta}_{\max}\)
        \Require Initial posterior correction
        \(\Delta\boldsymbol{\theta}^{+}_{0}=\mathbf{0}\)
        and covariance \(\mathbf{P}^{+}_{0}\)
        \Require Buffer-level process and measurement covariances
        \(\mathbf{Q}_{B}\) and \(\mathbf{R}_{B}\)

        \Statex \textbf{Notation:}
        \((\cdot)^{-}\) and \((\cdot)^{+}\) denote the EKF prior and
        posterior, respectively.

        \State \(t\gets0\)

        \While{the vehicle is operating}
            \State
            \(\mathcal{D}_{t}\gets\emptyset,\quad
            \mathbf{Y}_{t}\gets\emptyset\)

            \For{\(k=t,\ldots,t+B-1\)}
                \State Obtain
                \(\mathbf{x}_{k},\mathbf{u}_{k},\mathbf{m}_{k}\)
                \label{line:collect_start}

                \State Apply \(\mathbf{u}_{k}\) and measure
                \(\mathbf{y}_{k+1}
                =
                [\mathbf{v}_{k+1},
                \boldsymbol{\omega}_{k+1}]\)

                \State Append
                \((\mathbf{x}_{k},\mathbf{u}_{k},\mathbf{m}_{k})\)
                to \(\mathcal{D}_{t}\) and
                \(\mathbf{y}_{k+1}\) to \(\mathbf{Y}_{t}\)
                \label{line:collect_end}
            \EndFor

            \State
            \(\Delta\boldsymbol{\theta}^{-}_{t+B}
            \gets
            \Delta\boldsymbol{\theta}^{+}_{t}\)
            \label{line:stack_buffer}

            \State
            \(\mathbf{P}^{-}_{t+B}
            \gets
            \mathbf{P}^{+}_{t}
            +
            \mathbf{Q}_{B}\)

            \State
            \(\hat{\mathbf{Y}}_{t}\gets\emptyset\)

            \For{each
                \((\mathbf{x}_{k},\mathbf{u}_{k},\mathbf{m}_{k})
                \in\mathcal{D}_{t}\)}
                \State
                \(\bar{\boldsymbol{\theta}}_{k}
                \gets
                h_{\theta}
                (E_{\mathrm{s}}(\mathbf{m}^{\mathrm{s}}_{k}))\)
                \label{line:stack_pred}

                \State
                \(\boldsymbol{\theta}^{-}_{k}
                \gets
                \operatorname{clip}\!\left(
                    \bar{\boldsymbol{\theta}}_{k}
                    +
                    \Delta\boldsymbol{\theta}^{-}_{t+B},
                    \boldsymbol{\theta}_{\min},
                    \boldsymbol{\theta}_{\max}
                \right)\)

                \State
                \(\hat{\mathbf{y}}_{k+1}
                \gets
                \mathcal{V}\!\left(
                    F_{\boldsymbol{\Phi}}\!\left(
                        \mathbf{x}_{k},
                        \mathbf{u}_{k},
                        \mathbf{m}_{k};
                        \boldsymbol{\theta}^{-}_{k}
                    \right)
                \right)\)

                \State Append \(\hat{\mathbf{y}}_{k+1}\) to
                \(\hat{\mathbf{Y}}_{t}\)
                \label{line:end_pred}
            \EndFor

            \State
            \(\mathbf{e}_{t+B}
            \gets
            \mathbf{Y}_{t}
            -
            \hat{\mathbf{Y}}_{t}\)

            \State
            \(\mathbf{H}_{t+B}
            \gets
            \left.
            \dfrac{
                \partial\hat{\mathbf{Y}}_{t}
            }{
                \partial\Delta\boldsymbol{\theta}
            }
            \right|_{
                \Delta\boldsymbol{\theta}^{-}_{t+B}
            }\)
            \Comment{Buffered measurement Jacobian}

            \State
            \(\mathbf{S}_{t+B}
            \gets
            \mathbf{H}_{t+B}
            \mathbf{P}^{-}_{t+B}
            \mathbf{H}_{t+B}^{\top}
            +
            \mathbf{R}_{B}\)
            \Comment{Innovation covariance}

            \State
            \(\mathbf{K}_{t+B}
            \gets
            \mathbf{P}^{-}_{t+B}
            \mathbf{H}_{t+B}^{\top}
            \mathbf{S}_{t+B}^{-1}\)
            \Comment{Kalman gain}

            \State
            \(\Delta\boldsymbol{\theta}^{+}_{t+B}
            \gets
            \Delta\boldsymbol{\theta}^{-}_{t+B}
            +
            \mathbf{K}_{t+B}
            \mathbf{e}_{t+B}\)

            \State
            \(\mathbf{P}^{+}_{t+B}
            \gets
            \left(
                \mathbf{I}
                -
                \mathbf{K}_{t+B}
                \mathbf{H}_{t+B}
            \right)
            \mathbf{P}^{-}_{t+B}\)
            \label{line:ekf_update}

            \State \(t\gets t+B\)
            \Comment{Start a new buffer}
            \label{line:restart_buffer}
        \EndWhile
    \end{algorithmic}
\end{algorithm}

\section{Implementation}
\label{sec:implementation}

\subsection{Datasets and Preprocessing}
\label{sec:implementation_data}

We train and evaluate NeSAM using simulated and physical
vehicle-terrain interaction data. The simulation dataset is collected
in Verti-Bench~\cite{vertibench2025} over geometrically, semantically,
and physically diverse deformable terrain. We collect \(500\)
trajectories, each approximately \(25\,\mathrm{s}\) long, resulting in
approximately \(3.5\,\mathrm{h}\) of simulated interaction data. At the
\(10\,\mathrm{Hz}\) sampling rate, this corresponds to approximately
\(125{,}000\) transitions. The physical dataset is collected using the
Verti-4-Wheeler~\cite{datar2024toward} platform on a scaled uneven off-road testbed
Verti-Arena~\cite{chen2025verti}. We collect \(50\) trajectories, each
approximately \(30\,\mathrm{s}\) long, resulting in approximately
\(25\,\mathrm{min}\) of physical interaction data, or \(15{,}000\)
transitions. For both datasets, trajectories are split at the trajectory
level using an \(80\%/20\%\) train/test split to prevent transitions
from the same trajectory from appearing in both sets.

During data collection, the vehicle performs randomized sinusoidal
exploration to generate diverse steering curvatures and speed profiles.
The steering command follows
\(u_{\mathrm{steer}}(t)=\sin(\omega_{\mathrm{s}}t)\), where
\(\omega_{\mathrm{s}}\sim\mathcal{U}(0.1,0.5)\,\mathrm{Hz}\).
The commanded speed is
\(u_{\mathrm{speed}}(t)=v_{\mathrm{c}}+
A\sin(\omega_{\mathrm{v}}t)\), where
\(\omega_{\mathrm{v}}\sim\mathcal{U}(0.1,2.5)\,\mathrm{Hz}\).
We sample
\(v_{\min}\sim\mathcal{U}(1,2)\,\mathrm{m/s}\) and
\(v_{\max}\sim\mathcal{U}(3,4)\,\mathrm{m/s}\), with
\(A=(v_{\max}-v_{\min})/2\) and
\(v_{\mathrm{c}}=(v_{\max}+v_{\min})/2\).
This exploration exposes the model to varying vehicle velocities,
turning rates, wheel slip, and wheel-terrain interaction conditions.

To introduce variation in deformable-soil properties in Verti-Bench,
each semantic terrain class is associated with a bounded distribution
of terramechanics parameters rather than a single fixed parameter
vector. For each terrain region \(r\), a soil parameter vector
\(\boldsymbol{\theta}^{(r)}\) is independently sampled from the
distribution associated with its semantic class. The sampled parameters
remain fixed within the region but vary across regions and environments.
Consequently, regions sharing the same semantic label can exhibit
different pressure-sinkage and shear responses, preventing the
semantic observation from uniquely determining the underlying soil
properties.

For each transition, we record
\((\mathbf{x}_t,\mathbf{u}_t,\mathbf{x}_{t+1})\) together with
vehicle-aligned \(128\times128\) elevation and RGB semantic patches.
All observations are synchronized at \(10\,\mathrm{Hz}\), corresponding
to a fixed sampling interval of \(\Delta t=0.1\,\mathrm{s}\).
Elevation patches are globally normalized to \([-1,1]\), while the RGB
semantic patches are normalized channel-wise to the same range.
Further simulation and physical evaluation protocols are described in
Sec.~\ref{sec:experiments}.

\subsection{Terrain Representation and Soil Prediction}
\label{sec:implementation_terrain}

We use separate U-Net autoencoders for elevation and semantic terrain
observations. The elevation autoencoder accepts a single-channel input,
whereas the semantic autoencoder accepts a three-channel RGB input.
Except for their modality-specific input layers, both use the same
encoder architecture and produce an
\(18\times16\times16\) spatial feature map. The two autoencoders are
pretrained independently using mean-squared reconstruction loss and
Adam with a learning rate of \(10^{-4}\) and a batch size of \(64\).
After pretraining, the decoder branches are discarded and the terrain
encoders remain frozen during kinodynamic-model training.

The semantic representation is additionally processed by the soil
predictor \(h_{\theta}\). The predictor consists of two convolutional
layers with channel dimensions
\(18\!\rightarrow\!32\!\rightarrow\!64\), followed by a
\(64\!\rightarrow\!64\!\rightarrow\!6\) multilayer perceptron with
SiLU activations after the hidden layers. Its six-dimensional output
provides the nominal soil estimate
\(\bar{\boldsymbol{\theta}}_t\) used by the differentiable
terramechanics model. The soil predictor contains \(28{,}262\)
trainable parameters and is optimized jointly with the kinodynamic
predictor.

\subsection{Neuro-Symbolic Predictor and Training}
\label{sec:implementation_predictor}

The interaction tokens defined in Sec.~\ref{sec:method} are linearly
projected to a \(128\)-dimensional hidden representation and processed
by a causal Transformer. The Transformer contains two pre-normalized
self-attention blocks with four attention heads. Each block uses a
feed-forward network with dimensions
\(128\!\rightarrow\!256\!\rightarrow\!128\) and GELU activation.
Positional embeddings support a maximum sequence length of \(64\)
tokens, while the effective interaction history is \(32\) tokens,
corresponding to \(3.2\,\mathrm{s}\) at \(10\,\mathrm{Hz}\).
A key-value cache is used during sequential prediction to reuse the
representations of previous interaction tokens.

The final Transformer representation is passed to the prediction heads
used by the neuro-symbolic transition model. For the four-wheeled
platform, the wheel-interaction prediction includes the wheel-level
quantities required by the differentiable terramechanics model, while
the state-residual head predicts the \(12\)-dimensional chassis
correction \(\Delta\mathbf{x}^{\mathrm{res}}_t\). In particular, the
predicted wheel sinkages
\(\hat{\mathbf{s}}_t\in\mathbb{R}^{4}\) represent the deformation of
the terrain beneath the four wheels. These intermediate physical
quantities receive no direct supervision and are learned through the
vehicle-state prediction objective.

NeSAM is trained for \(40\) epochs using Adam, with \(250\)
mini-batches per epoch and a batch size of \(32\). The learned
kinodynamic components are optimized with a learning rate of
\(10^{-4}\). Training uses fully autoregressive \(K=32\)-step
rollouts, corresponding to a \(3.2\,\mathrm{s}\) prediction horizon at
\(10\,\mathrm{Hz}\). The objective is the mean-squared error between
the predicted and ground-truth vehicle states over the rollout horizon.
During training, gradients are propagated through the learned
interaction predictor, differentiable terramechanics, Newton-Euler
kinodynamics, and state-residual branch, while the pretrained terrain
encoders remain fixed.

\section{Experiments}
\label{sec:experiments}

We evaluate NeSAM from two complementary aspects. First, we evaluate
long-horizon kinodynamic modeling accuracy against learning-based
state-of-the-art baselines in both simulation and physical experiments.
Second, we evaluate the effect of NeSAM's online soil adaptation in
closed-loop trajectory tracking by comparing the same NeSAM model with
and without EKF-based soil adaptation. The latter therefore serves as
an adaptation ablation rather than a comparison against other
kinodynamic baselines. Experiments are conducted in the
Verti-Bench simulator~\cite{vertibench2025} and on a physical
off-road testbed Verti-Arena~\cite{chen2025verti}.

For prediction evaluation, we use the held-out \(20\%\) test split
described in Sec.~\ref{sec:implementation_data}. Prediction performance
is evaluated using fully autoregressive \(K=32\)-step rollouts,
corresponding to \(3.2\,\mathrm{s}\) at \(10\,\mathrm{Hz}\), with
mean absolute error (MAE) over position \((x,y,z)\) and orientation
\((\mathrm{roll},\mathrm{pitch},\mathrm{yaw})\).

We compare NeSAM against a pure end-to-end Transformer predictor and
TAL~\cite{datar2024terrain} using elevation and semantic observations
jointly (TAL Elev.+Sem.) or either modality alone. We further evaluate
two NeSAM ablations: \emph{NeSAM w/o SR}, which removes the learned
state-residual branch, and \emph{NeSAM, Sinkage \(=0\)}, which
suppresses the predicted wheel sinkage while retaining the remaining
neuro-symbolic pipeline.

\subsection{Simulation Experiments in Verti-Bench}
\label{sec:sim_experiments}

We first evaluate NeSAM in Verti-Bench using the simulation dataset
described in Sec.~\ref{sec:implementation_data}.
For kinodynamic prediction, each rollout is initialized from the
measured vehicle state and subsequently propagated using the model's own
predictions, with terrain observations queried at the predicted vehicle
poses.

For closed-loop trajectory tracking, NeSAM is integrated with MPPI and
evaluated under out-of-distribution (OOD) soil conditions. We construct
the OOD setting by evaluating the vehicle on soil parameter
configurations outside the distributions used for model training,
thereby introducing a mismatch between the nominal semantic soil
estimate and the soil properties governing the simulated
wheel-terrain interaction.

\subsubsection{Long-Horizon Kinodynamic Prediction}
\label{sec:sim_prediction}

As shown in Table~\ref{tab:sim_prediction_results}, NeSAM achieves the
lowest error across all six reported state dimensions. Compared with
the Transformer, NeSAM reduces the \(x\)-, \(y\)-, and \(z\)-errors
from \(0.89\), \(0.90\), and \(0.09\,\mathrm{m}\) to \(0.80\),
\(0.78\), and \(0.06\,\mathrm{m}\), respectively. Yaw error decreases
from \(6.76^\circ\) to \(4.18^\circ\).

The TAL variants exhibit larger translational errors despite
conditioning on elevation and/or semantic terrain information. This
indicates that terrain conditioning alone provides limited ability to
capture deformation-dependent wheel-terrain interaction over long
autoregressive horizons. NeSAM instead couples learned
terrain-conditioned interaction prediction with explicit
terramechanics and Newton-Euler propagation.

The ablations further demonstrate the contributions of the
neuro-symbolic components. Removing the state residual increases the
\(x\)- and \(y\)-errors from \(0.80\) and \(0.78\,\mathrm{m}\) to
\(0.99\) and \(0.98\,\mathrm{m}\), respectively, and degrades the
overall prediction performance. Suppressing wheel sinkage also
increases error across all six dimensions. Relative to the
Sinkage \(=0\) variant, full NeSAM reduces \(x\)-, \(y\)-, and
\(z\)-errors by approximately \(7.1\%\), \(7.8\%\), and \(18.3\%\),
respectively. These results support the complementary roles of the
learned state correction and deformation-related wheel variables.

\subsubsection{Trajectory Tracking with Online Soil Adaptation}
\label{sec:sim_tracking}

We next isolate the effect of online soil adaptation on closed-loop
trajectory tracking. Unlike the prediction experiment above, this
experiment does not compare against external kinodynamic baselines.
Instead, the same trained NeSAM model and MPPI controller are used in
both configurations, with EKF-based soil adaptation being the only
difference. The \emph{Fixed Soil} configuration retains the nominal
soil estimate, whereas the \emph{Adapted Soil} configuration refines
the soil parameters online from observed vehicle motion.

Each configuration is evaluated over five runs on the same reference
trajectory under the OOD soil setting. We report trajectory completion,
traversal time for completed runs, Hausdorff distance (HD) between the
executed and reference trajectories, and mean absolute roll and pitch.

\begin{table}[t]
\centering
\caption{Autoregressive \(32\)-step kinodynamic prediction in
Verti-Bench.}
\label{tab:sim_prediction_results}
\scriptsize
\renewcommand{\arraystretch}{1.4}
\setlength{\tabcolsep}{5pt}
\begin{tabular}{lcccccc}
\hline
\textbf{Model}
& \(\mathbf{x}\downarrow\)
& \(\mathbf{y}\downarrow\)
& \(\mathbf{z}\downarrow\)
& \textbf{Roll}\(\downarrow\)
& \textbf{Pitch}\(\downarrow\)
& \textbf{Yaw}\(\downarrow\) \\
\hline
Transformer
& 0.89m & 0.90m & 0.09m & 2.64\degree & 2.01\degree & 6.76\degree \\

TAL Elev.+Sem.
& 1.19m & 1.08m & 0.12m & 2.64\degree & 2.24\degree & 6.07\degree \\

TAL Elev.
& 1.23m & 1.08m & 0.13m & 2.75\degree & 2.12\degree & 6.19\degree \\

TAL Sem.
& 1.34m & 1.11m & 0.15m & 2.69\degree & 2.18\degree & 6.99\degree \\
\hline
NeSAM w/o SR
& 0.99m & 0.98m & 0.10m & 2.58\degree & 2.29\degree & 7.05\degree \\

NeSAM, Sinkage \(=0\)
& 0.86m & 0.84m & 0.07m & 2.81\degree & 2.24\degree & 4.41\degree \\

\rowcolor{gray!15}
\textbf{NeSAM}
& \textbf{0.80m}
& \textbf{0.78m}
& \textbf{0.06m}
& \textbf{2.35\degree}
& \textbf{1.99\degree}
& \textbf{4.18\degree} \\
\hline
\end{tabular}
\end{table}

\begin{table}[h]
\centering
\caption{Trajectory-tracking performance in Verti-Bench under OOD
soil conditions.}
\label{tab:sim_tracking_results}
\scriptsize
\renewcommand{\arraystretch}{1.2}
\setlength{\tabcolsep}{8pt}
\begin{tabular}{lcc}
\hline
\textbf{Metric}
& \textbf{Adapted Soil}
& \textbf{Fixed Soil} \\
\hline
Completion \(\uparrow\)
& \textbf{4/5}
& 0/5 \\

Traversal Time \(\downarrow\)
& 100.0 $\pm$ 6.5
& - \\

HD \(\downarrow\)
& \textbf{2.10m $\pm$ 1.25m}
& 6.87m $\pm$ 3.12m \\

Roll \(\downarrow\)
& \textbf{4.56\degree $\pm$ 5.77\degree}
& 6.72\degree $\pm$ 5.17\degree \\

Pitch \(\downarrow\)
& \textbf{6.59\degree $\pm$ 8.26\degree}
& 8.03\degree $\pm$ 9.23\degree \\
\hline
\end{tabular}
\end{table}

As shown in Table~\ref{tab:sim_tracking_results}, none of the five
fixed-soil runs completes the reference trajectory, whereas online
adaptation achieves \(4/5\) completions. The Hausdorff distance
decreases from \(6.87\) to \(2.10\,\mathrm{m}\), corresponding to a
\(69.4\%\) reduction in geometric tracking error. Mean absolute roll
and pitch also decrease under online adaptation. Because no fixed-soil
trial completes the trajectory, traversal time cannot be meaningfully
compared between the two configurations. These results show that
refining the soil parameters online substantially improves closed-loop
tracking when the nominal soil model does not match the encountered
terrain.

\subsection{Physical Experiments on Verti-Arena}
\label{sec:physical_experiments}

\begin{figure}[t]
    \centering
    \includegraphics[width=\columnwidth]{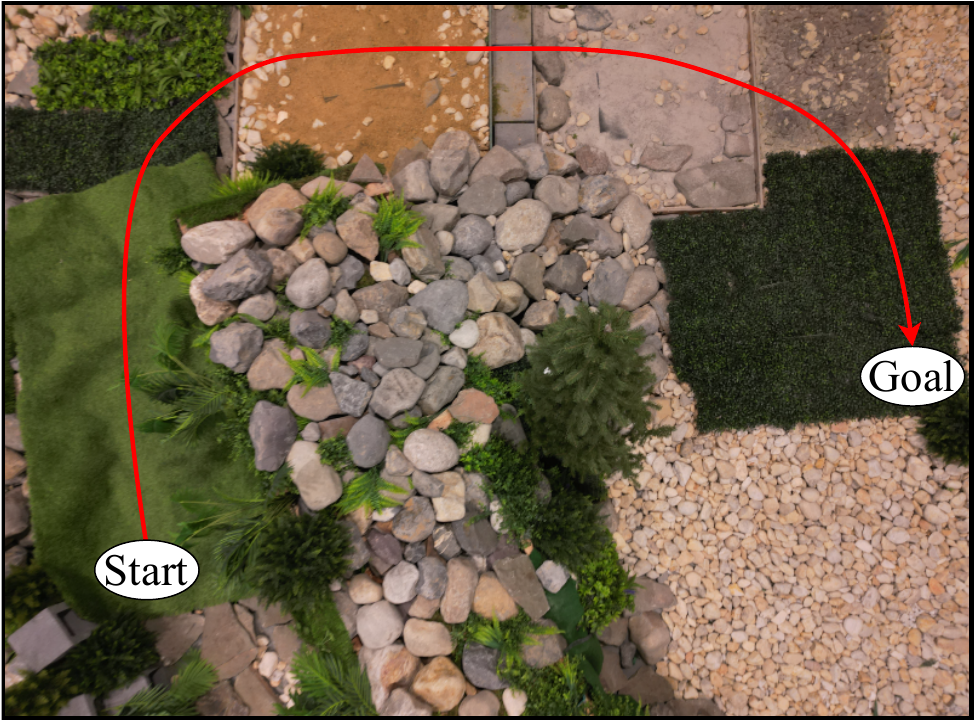}
    \caption{Verti-Arena and reference trajectory used
    for evaluating NeSAM.}
    \label{fig:physical_testbed}
\end{figure}

We further evaluate NeSAM using the Verti-4-Wheeler~\cite{datar2024toward} platform on Verti-Arena. The Verti-Arena
contains terrain regions with different deformation characteristics,
including grass, dirt, and deformable sand, together with uneven geometric
structures that induce variations in wheel-terrain interaction and
vehicle attitude. In particular, the reference trajectory traverses a
grass-to-sand transition from a predefined start position to the goal,
as illustrated in Fig.~\ref{fig:physical_testbed}.

The physical prediction evaluation follows the same \(32\)-step
autoregressive protocol used in simulation and is performed on the physical test trajectories. For closed-loop trajectory
tracking, the same trained NeSAM predictor and MPPI controller are
evaluated with fixed and online-adapted soil estimates over five trials
for each configuration.

\subsubsection{Long-Horizon Kinodynamic Prediction}
\label{sec:physical_prediction}

Table~\ref{tab:physical_prediction_results} shows that the prediction
advantages of NeSAM transfer to physical vehicle trajectories. NeSAM
achieves the lowest error in five of the six state dimensions,
including all three position components, roll, and yaw. TAL Elev.+Sem.
achieves the lowest pitch error of \(3.61^\circ\). Compared with the Transformer baseline, NeSAM reduces the \(x\)-,
\(y\)-, and \(z\)-errors from \(0.082\), \(0.070\), and
\(0.019\,\mathrm{m}\) to \(0.068\), \(0.059\), and
\(0.012\,\mathrm{m}\), corresponding to reductions of \(17.1\%\),
\(15.7\%\), and \(36.8\%\), respectively. Roll, pitch, and yaw errors
are reduced from \(3.78^\circ\), \(5.21^\circ\), and \(7.05^\circ\)
to \(3.15^\circ\), \(4.18^\circ\), and \(5.79^\circ\), respectively.

\begin{table}[h!]
\centering
\caption{Autoregressive \(32\)-step kinodynamic prediction on physical
testbed.}
\label{tab:physical_prediction_results}
\scriptsize
\renewcommand{\arraystretch}{1.4}
\setlength{\tabcolsep}{4.4pt}
\begin{tabular}{lcccccc}
\hline
\textbf{Model}
& \(\mathbf{x}\downarrow\)
& \(\mathbf{y}\downarrow\)
& \(\mathbf{z}\downarrow\)
& \textbf{Roll}\(\downarrow\)
& \textbf{Pitch}\(\downarrow\)
& \textbf{Yaw}\(\downarrow\) \\
\hline
Transformer
& 0.082m & 0.070m & 0.019m & 3.78\degree & 5.21\degree & 7.05\degree \\

TAL Elev.+Sem.
& 0.083m & 0.072m & 0.021m & 3.55\degree & \textbf{3.61\degree} & 6.59\degree \\

TAL Elev.
& 0.089m & 0.095m & 0.022m & 3.49\degree & 4.24\degree & 7.79\degree \\

TAL Sem.
& 0.111m & 0.092m & 0.028m & 3.61\degree & 3.67\degree & 7.05\degree \\
\hline
NeSAM w/o SR
& 0.177m & 0.151m & 0.031m & 3.89\degree & 5.33\degree & 7.74\degree \\

NeSAM, Sinkage \(=0\)
& 0.077m & 0.068m & 0.015m & 3.44\degree & 4.64\degree & 6.13\degree \\

\rowcolor{gray!15}
\textbf{NeSAM}
& \textbf{0.068m}
& \textbf{0.059m}
& \textbf{0.012m}
& \textbf{3.15\degree}
& 4.18\degree
& \textbf{5.79\degree} \\
\hline
\end{tabular}
\end{table}

The physical ablations exhibit the same overall trend as the simulation
results. Removing the state residual substantially increases the
translational errors and degrades all six state metrics. Suppressing
wheel sinkage also consistently increases prediction error. Relative
to the Sinkage \(=0\) variant, full NeSAM reduces the \(x\)-, \(y\)-,
and \(z\)-errors by \(11.7\%\), \(13.2\%\), and \(20.0\%\),
respectively. The consistent benefit across simulation and physical
data supports the complementary roles of the learned state correction
and deformation-aware wheel-terrain interaction.

\subsubsection{Trajectory Tracking with Online Soil Adaptation}
\label{sec:physical_tracking}

Finally, we isolate the effect of online soil adaptation during
physical closed-loop trajectory tracking. As in simulation, both
configurations use the same trained NeSAM model, MPPI controller, and
reference trajectory; the only difference is whether EKF-based soil
adaptation is enabled. The \emph{Fixed Soil} configuration retains the
nominal soil estimate, whereas the \emph{Adapted Soil} configuration
updates the soil parameters online using measured vehicle motion.

We perform five trials for each configuration. In addition to
trajectory completion and traversal time, we compute the mean absolute
roll and pitch for each trial and report their mean and standard
deviation across all five trials.

\begin{table}[h]
\centering
\caption{Trajectory-tracking performance on Verti-Arena.}
\label{tab:physical_tracking_results}
\scriptsize
\renewcommand{\arraystretch}{1.2}
\setlength{\tabcolsep}{8pt}
\begin{tabular}{lcc}
\hline
\textbf{Metric}
& \textbf{Adapted Soil}
& \textbf{Fixed Soil} \\
\hline
Completion \(\uparrow\)
& \textbf{3/5}
& 1/5 \\

Traversal Time \(\downarrow\)
& 21.33s $\pm$ 1.39s
& 21.50s \\

Roll \(\downarrow\)
& \textbf{7.27\degree $\pm$ 3.71\degree}
& 9.23\degree $\pm$ 5.22\degree \\

Pitch \(\downarrow\)
& \textbf{8.01\degree $\pm$ 4.49\degree}
& 10.40\degree $\pm$ 4.56\degree \\
\hline
\end{tabular}
\end{table}

As shown in Table~\ref{tab:physical_tracking_results}, online
adaptation increases trajectory completion from \(1/5\) to \(3/5\).
Across all five trials, mean absolute roll decreases from
\(9.23^\circ\) to \(7.27^\circ\), while mean absolute pitch decreases
from \(10.40^\circ\) to \(8.01^\circ\), corresponding to reductions
of \(21.2\%\) and \(23.0\%\), respectively. The three completed adapted
trials have a mean traversal time of \(21.33\pm1.39\,\mathrm{s}\),
whereas the single completed fixed-soil trial requires
\(21.50\,\mathrm{s}\). Because only one fixed-soil trial completes the
trajectory, we do not draw a quantitative conclusion from traversal
time.

Together with the OOD simulation results, the physical experiments
show that adapting the interpretable soil parameters enables NeSAM to
better accommodate mismatch between the nominal terramechanics model
and the encountered wheel-terrain interaction, resulting in more
reliable closed-loop trajectory tracking.
\section{Conclusion}
\label{sec:conclusion}
We present NeSAM, a neuro-symbolic kinodynamic modeling framework for
off-road mobility over deformable terrain. By combining learned terrain
representations, prediction of physically meaningful wheel-terrain
interaction variables, differentiable Bekker-Wong terramechanics, and
a learned state residual, NeSAM improves long-horizon vehicle-motion
prediction in both simulation and physical experiments. When integrated
with MPPI, the EKF-based online soil adaptation further improves
closed-loop trajectory tracking by refining the soil parameters from
observed vehicle motion as terrain conditions change.

A current limitation is that NeSAM assumes wheel-terrain interaction
can be adequately represented by the underlying Bekker-Wong
terramechanics formulation. Although online adaptation can refine the
associated soil parameters, it cannot modify the constitutive
pressure-sinkage and shear relationships themselves. Consequently,
terrain behaviors that deviate substantially from these assumptions may
remain difficult to model even after parameter adaptation. Future work
will investigate richer differentiable terrain models and learned
physics corrections that can relax these structural assumptions while
retaining the interpretability and adaptability of the neuro-symbolic
framework.



\bibliographystyle{IEEEtran}
\bibliography{IEEEabrv,references}

\end{document}